%% file: main.tex
\def\year{2021}\relax
\documentclass[letterpaper]{article} 
\usepackage{aaai21}  
\usepackage{times}  
\usepackage{helvet} 
\usepackage{courier}  
\usepackage[hyphens]{url}  
\usepackage{graphicx} 

\usepackage{adjustbox}
\usepackage{booktabs}
\usepackage{multirow}
\usepackage[dvipsnames]{xcolor}
\usepackage{amsmath}
\usepackage{amsfonts} 
\usepackage{algorithmicx}
\usepackage{algorithm}
\usepackage{algpseudocode}
\usepackage{graphicx} 
\usepackage{enumitem}
\usepackage{hyperref}
\usepackage[switch]{lineno}

\usepackage{natbib}  
\usepackage{caption} 
\title{\textsc{Filter}: An Enhanced Fusion Method for Cross-lingual Language Understanding}
\author{
    Yuwei Fang\footnote{Equal Contribution}, Shuohang Wang\footnotemark[1], Zhe Gan, Siqi Sun, Jingjing Liu\\
}

\affiliations{
    Microsoft Dynamics 365 AI Research\\
    {\tt \small{ \{yuwfan,shuohang.wang,zhe.gan,siqi.sun,jingjl\}@microsoft.com}}\\
}

\begin{document}
\maketitle

\begin{abstract}
Large-scale cross-lingual language models (LM), such as mBERT, Unicoder and XLM, have achieved great success in cross-lingual representation learning. However, when applied to zero-shot cross-lingual transfer tasks, most existing methods use only single-language input for LM finetuning, without leveraging the intrinsic cross-lingual alignment between different languages that proves essential for multilingual tasks. In this paper, 
we propose \textsc{Filter}, an enhanced fusion method that takes cross-lingual data as input
for XLM finetuning. Specifically, \textsc{Filter} first encodes text input in the source language and its translation in the target language independently in the shallow layers, then performs cross-language fusion to extract multilingual knowledge in the intermediate layers, and finally performs further language-specific encoding. During inference, the model makes predictions based on the text input in the target language and its translation in the source language.
For simple tasks such as classification, translated text in the target language shares the same label as the source language. However, this shared label becomes less accurate or even unavailable for more complex tasks such as question answering, NER and POS tagging. To tackle this issue, we further propose an additional KL-divergence \emph{self-teaching} loss for model training, based on auto-generated \emph{soft} pseudo-labels for translated text in the target language.  
Extensive experiments demonstrate that \textsc{Filter} achieves new state of the art on two challenging multilingual multi-task benchmarks, XTREME and XGLUE.\footnote{Our code is released at \url{https://github.com/yuwfan/FILTER}.}
\end{abstract}
\input{intro}
\input{related}
\input{model}

\input{exp}
\bibliography{main}
\bibliographystyle{aaai21}

\clearpage
\input{appendix}

\end{document}

%% file: intro.tex
\section{Introduction}

Cross-lingual low-resource adaptation has been a critical and exigent problem in the NLP field, despite recent success in large-scale language models (mostly trained on English with abundant training corpora). How to adapt models trained in high-resource languages (\emph{e.g.}, English) to low-resource ones (most of the 6,900 languages in the world) still remains challenging.
To address the proverbial domain gap between languages, three schools of approach have been widely studied. ($i$) \emph{Unsupervised pre-training}: to learn a
universal encoder (cross-lingual language model) for different languages.
For example, mBERT~\cite{devlin-etal-2019-bert}, Unicoder~\cite{huang2019unicoder} and XLM~\cite{lample2019cross} have achieved strong performance on many cross-lingual tasks by successfully transferring knowledge from source language to a target one. 
($ii$) \emph{Supervised training}: to enforce models insensitive to labeled data across different languages, through teacher forcing~\cite{wu2020single} or adversarial learning~\cite{cao2020jointly}. 
($iii$) \emph{Translation}: to translate either source language to the target one, or vice versa ~\cite{cui2019cross,hu2020xtreme,liang2020xglue}, so that training and inference can be performed in the same language. 

The translation approach has proven highly effective on recent multilingual benchmarks. For example, the \emph{translate-train} method
has achieved state of the art on XTREME~\cite{hu2020xtreme} and XGLUE~\cite{liang2020xglue}.
However, translate-train is simple data augmentation, which doubles training data by translating source text into target languages. Thus, only single-language input is considered for finetuning with augmented data, leaving out cross-lingual alignment between languages unexplored.
Dual BERT~\cite{cui2019cross} is recently proposed to make use of the representations learned from source language to help target language understanding. 
However, it only injects information from the source language into the decoder of target language, without scoping into the intrinsic relations between languages. 

\begin{figure*}[t]
\centering
\includegraphics[width=7in]{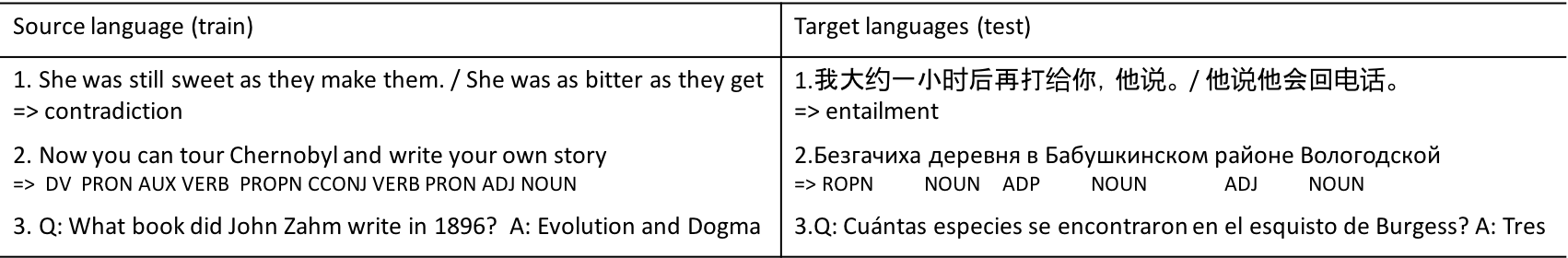}
\caption{Examples from XTREME for cross-lingual natural language inference, part-of-speech tagging, and question answering tasks. The source language is English; the target language can be any other languages.}
\label{fig:xtreme_examples}
\end{figure*}

Motivated by this, we propose \textsc{Filter},\footnote{\textbf{F}usion in the \textbf{I}ntermediate \textbf{L}ayers of \textbf{T}ransform\textbf{ER}} a generic and flexible framework that leverages translated data to enforce fusion between languages for better cross-lingual language understanding. As illustrated in Figure~\ref{fig:model}(c),
\textsc{Filter} first ($i$) encodes a translated language pair separately in shallow layers; then ($ii$) performs cross-lingual fusion between languages in the intermediate layers; and finally ($iii$) encodes language-specific representations in deeper layers. Compared to the translate-train baseline (Figure~\ref{fig:model}(a)), \textsc{Filter} learns additional cross-lingual alignment that is instrumental to cross-lingual representations. Furthermore, compared to simply concatenating the language pair as the input of XLM (Figure~\ref{fig:model}(b)), \textsc{Filter} strikes a well-measured balance between cross-lingual fusion and individual language representation learning. 

For classification tasks such as natural language inference, translated text in the target language shares the same label as the source language. However, for question answering (QA) tasks, the answer span in the translated text of target language generally differs from that in the source language. For sequential labeling tasks such as NER (Named Entity Recognition) and POS (Part-of-Speech) tagging,
the sequence of labels in the target language becomes unavailable, as the linguistic structure of sentences greatly varies across different languages.  
To bridge the gap, we propose to generate \emph{soft} pseudo-labels for translated text, and use an additional KL-divergence \emph{self-teaching} loss for model training. 
Specifically, we first train a teacher \textsc{Filter} model, to collect the inference probabilities for the translated text of all training samples, which will be used as pseudo soft-labels to train a student \textsc{Filter} as the final prediction model. For QA, POS and NER tasks, this self-training process generates more reliable and accurate labels than hard label assignment on translated text, leading to better model performance. For classification tasks where the target label is identical to the source, self-teaching loss proves to also improve performance, by serving as an effective regularizer.    

The main contributions are summarized as follows.
($i$) We propose \textsc{Filter}, a new approach to cross-lingual language understanding by leveraging intrinsic linguistic alignment between languages for XLM finetuning.
($ii$) We propose a self-teaching loss to address the unreliable/unavailable label issue in target language, boosting model performance across diverse NLP tasks. 
($iii$) We achieve Top-1 performance on both XTREME and XGLUE benchmarks, outperforming previous state of the art by absolute 8.8 and 2.2 points (published and unpublished) in XTREME, and 4.0 points in XGLUE, respectively.

%% file: related.tex
\section{Related Work}
\paragraph{Cross-lingual Datasets}
Cross-lingual language understanding has been investigated for many NLP tasks, where the knowledge learned from a pivot language (\emph{e.g.}, English) is transferred to other languages indirectly, as labeled data in low-resource languages are often scarce.
There exist many multilingual corpora for diverse NLP tasks. \citet{nivre2016universal} released a collection of multilingual treebanks on universal dependencies for 33 languages.
\citet{pan2017cross} introduced cross-lingual name tagging and linking for 282 languages.
Other multilingual datasets range over tasks such as document classification, natural language inference, information retrieval, paraphrase identification, and summarization \cite{klementiev2012inducing, cer2017semeval, conneau2018xnli, sasaki2018cross, yang2019paws, zhu2019ncls}.

More recent studies on open-domain question answering and machine reading comprehension also introduced cross-lingual datasets, such as 
MLQA~\cite{lewis2019mlqa}, XQuAD~\cite{Artetxe:etal:2019}, and TyDiQA~\cite{clark2020tydi}.
Most recently, XTREME~\cite{hu2020xtreme} and XGLUE~\cite{liang2020xglue} released several datasets across multiple tasks, and set up public leaderboards for evaluating cross-lingual models.
In this paper, we work on both XTREME (see Figure~\ref{fig:xtreme_examples} for examples) and XGLUE to demonstrate the effectiveness of our proposed method.

\paragraph{Cross-lingual Models}
Most previous work tackles cross-lingual problems in two fashions: ($i$) cross-lingual zero-shot transfer; and ($ii$) translate-train/test. For cross-lingual zero-shot transfer, models are trained on labeled data in the source language only, and directly evaluated on target languages.
Early work focused on training multilingual word embeddings~\cite{mikolov2013exploiting,faruqui2014improving,xu2018unsupervised}, 
while more recent work proposed to
pre-train cross-lingual language models, such as mBERT~\cite{devlin-etal-2019-bert}, XLM~\cite{lample2019cross} and XLM-Roberta~\cite{conneau2019unsupervised}, to learn contextualized representations. 

\begin{figure*}[t]
\centering
\includegraphics[width=6.7in]{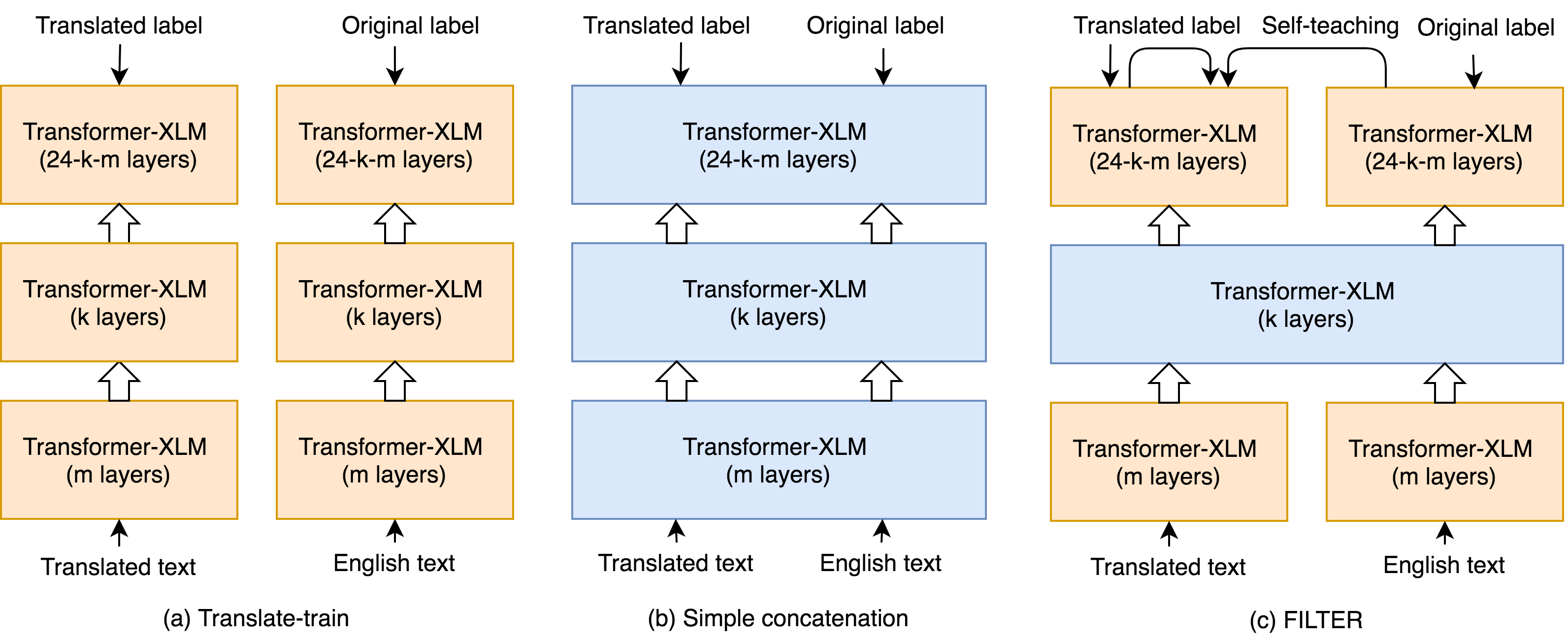}
\caption{Comparison between different methods for finetuning XLM-R model for the XTREME benchmark. (a) Translate-train baseline. (b) Another baseline via simple concatenation of translated text. (c) Proposed \textsc{Filter} approach. (a) and (b) can be considered as special instantiations of \textsc{Filter} by setting $m=24, k=0$ and $m=0, k=24$, respectively.}
\label{fig:model}
\end{figure*}

For translate-train/test, external 
machine translation tools are leveraged. 
A common approach is to augment training data by first translating all data in the source language to target languages, then train the model on translated data~\cite{hu2020xtreme,liang2020xglue}.
Another approach is translate-test~\cite{hu2020xtreme} or round-trip translation~\cite{zhu2019ncls}, which translates the text in the test set of target languages into source language, so that all the models trained in the source language can be directly applied for inference, and the prediction can be translated back to the target language if needed.
To enhance these translation-based pipelines,
\citet{cui2019cross} proposed to simultaneously model text in both languages to enrich the learned language representations.
\citet{huang2019cross} proposed to use adversarial transfer to enhance low-resource name tagging.
And \citet{cao2020jointly} proposed to jointly learn the alignment and perform summarization across languages. 
\textsc{Filter} follows the translate-train line of thought, but provides a better way to encode text in both source and target languages simultaneously.

%% file: model.tex
\section{Proposed Approach} 
In this section, we first introduce the proposed $\textsc{Filter}$ model architecture, then describe the self-teaching loss for model enhancement.
An overview of the framework is illustrated in Figure~\ref{fig:model}.

\subsection{$\textsc{Filter}$ Architecture}


Although the domain gap between languages has been largely reduced by translate-train method, translated text 
may not succeed in keeping the semantic meaning and label of the original text unchanged, due to quality constraint of translation tools.
Furthermore, the source language and translated target language are usually encoded separately, without tapping into the cross-lingual relations among different languages.
Therefore, we propose to use language pairs as input, and fuse the learned representations between languages through intermediate network layers, so that the model can learn cross-lingual information that is instrumental to inference in different languages.

The proposed $\textsc{Filter}$ model
consists of three components: ($i$) ``local'' Transformer layers for encoding the input language pair independently; ($ii$) cross-lingual fusion layers for leveraging the context in different languages;
and ($iii$) deeper domain-specific Transformer layers to shift the focus back on individual languages, after injecting information from the other language.
For notation, $\mathbf{S}\in \mathbb{R}^{d\times l_s}$ and $\mathbf{T}\in \mathbb{R}^{d\times l_t}$ are denoted as the word embedding matrix for text input $S$ and $T$ in the source and target language, respectively.
If tasks involve pairwise data, $S$ is the concatenation of a sequence pair, such as the context and question in QA tasks.
$T$ is translated from $S$ via translation tools. $d$ is the word embedding dimension. $l_s$ and $l_t$ are the lengths of the text input $S$ and $T$, respectively. 
Formally,
\begin{eqnarray}
\nonumber
\mathbf{H}^{s}_{l} &=& \text{Transformer-XLM}_{local}(\mathbf{S})\,, \\
\nonumber
\mathbf{H}^{t}_{l} &=& \text{Transformer-XLM}_{local}(\mathbf{T})\,,
\end{eqnarray}
where the position embeddings are counted from 0 for both sequences, $\mathbf{H}^{s}_{l}\in \mathbb{R}^{d\times l_s}$ and $\mathbf{H}^{t}_{l}\in \mathbb{R}^{d\times l_t}$ are ``local'' representations of the sequence pair.
We set the number of layers in $\text{Transformer-XLM}_{local}$ as $m$, which can be tuned for solving different cross-lingual tasks.
The concatenation of the local representations from both languages, $\left[\mathbf{H}^{s}_{l}; \mathbf{H}^{t}_{l}\right]\in \mathbb{R}^{d\times (l_s+l_t)}$, is the input for the next layer to learn the fusion between different languages, as follows:
\begin{equation}
\left[ \mathbf{H}^{s}_{f}; \mathbf{H}^{t}_{f} \right] = \text{Transformer-XLM}_{fuse} (\left[\mathbf{H}^{s}_{l}; \mathbf{H}^{t}_{l}\right] )\,, 
\end{equation}
where $[\cdot; \cdot]$ denotes the concatenation of two matrices, $\mathbf{H}^{s}_{f}\in \mathbb{R}^{d\times l_s}$ and $\mathbf{H}^{t}_{f}\in \mathbb{R}^{d\times l_t}$ are the representations in corresponding languages.
We set the number of layers in $\text{Transformer-XLM}_{fuse}$ as $k$, which is another hyper-parameter to control the cross-lingual fusion degree.
As the final goal is to predict the label in one language, we limit the top layers specifically designed to encode the text in one language, so that not too much noise is introduced from translated text in other languages. Specifically,
\begin{eqnarray}
\mathbf{H}^{s}_{d} &=& \text{Transformer-XLM}_{domain}(\mathbf{H}^{s}_{f}), \\
\mathbf{H}^{t}_{d} &=& \text{Transformer-XLM}_{domain}(\mathbf{H}^{t}_{f}), 
\end{eqnarray}
where $\mathbf{H}^{s}_{d}\in \mathbb{R}^{d\times l_s}$ and $\mathbf{H}^{t}_{d}\in \mathbb{R}^{d\times l_t}$ are the final representations for prediction.

As demonstrated in Figure~\ref{fig:model}, $\textsc{Filter}$ is realized by stacking the three types of transformer layer on top of each other.
$\textsc{Filter}$ is a generic framework for solving multilingual tasks, where $k$ and $m$ can be flexibly set to different values depending on the task. For example, for classification tasks, a smaller $k$ is desired; while for question answering, a larger $k$ is needed for absorbing richer cross-lingual information (see Experiments for empirical evidence). Since we use XLM-R as the backbone in our framework,
the number of layers in $\text{Transformer-XLM}_{domain}$ is $24-k-m$.
When $m=24, k=0$, $\textsc{Filter}$ degenerates to the translate-train baseline (Figure~\ref{fig:model}(a)).
When $m=0,k=24$, $\textsc{Filter}$ reduces to another baseline that simply concatenates the text in different languages for XLM finetuning (Figure~\ref{fig:model}(b)).


$\textsc{Filter}$ also stacks a task-specific linear layer on top of $\mathbf{H}^{s}_{d}$ and $\mathbf{H}^{t}_{d}$ to compute the candidate probabilities and we simplify the whole framework as follows:
\begin{eqnarray}
\nonumber
\boldsymbol{p}^s, \boldsymbol{p}^t &=& \textsc{Filter} \, (\mathbf{S}, \mathbf{T}), \\
\nonumber
\mathcal{L}^{s} &=& \text{Loss}_{task} (\boldsymbol{p}^s,  l^{s} ), \\
\mathcal{L}^{t} &=& \text{Loss}_{task} (\boldsymbol{p}^t,  l^{t} ),
\label{eqn:logit}
\end{eqnarray}
where $\boldsymbol{p}^s$ and $\boldsymbol{p}^t$ are task-specific probability vectors over candidates, used to compute the final loss based on the labels $l^{s}$ and $l^{t}$ from source and target languages, respectively. As shown in Figure~\ref{fig:xtreme_examples}, for natural language inference, the label can be entailment/contradiction/neutral; for question answering, the label is  an answer span positions; for NER and POS tagging, the supervision becomes a sequence of labels.

\begin{algorithm}[t]
\begin{algorithmic}[1]
\State \# Teacher model training
\State \# $S, l^s$: text and label in the source language
\State \# $T, l^t$: text and label in the target language
\For{all $S, l^s$} 
    \State $T$ = Translation ($S$);
    \State $l_t$ = Transfer from $l_s$ if available;
    \State Train \textsc{Filter}$_{tea}$ with $(S, l^s)$ and $(T, l^t)$;
\EndFor
\\
\State \# Self-teaching, \emph{i.e.}, student model training
\For{all $S, l^s, T, l^t$} 
    \State $\boldsymbol{p}_{tea}^s, \boldsymbol{p}_{tea}^t$ = \textsc{Filter}$_{tea}$ $(S, T)$
    \State Train \textsc{Filter}$_{stu}$ with $(S, l^s)$, $(T, l^t)$ and $(T,\boldsymbol{p}_{tea}^t)$
\EndFor
\end{algorithmic}
\caption{\textsc{Filter} Training Procedure.}
\label{alg}
\end{algorithm}

\subsection{Self-Teaching Loss}
The teacher-student framework, or distillation loss~\cite{hinton2015distilling}, has been widely adopted in many areas. In this paper, we propose to add self-teaching loss for training $\textsc{Filter}$, and it can be readily adapted to all the cross-lingual tasks.
As transferring the labels in source language to the corresponding translated text may introduce noise due to the word order or even semantic meaning changes after translation, the additional \emph{self-teaching} loss is to bridge this gap.

The proposed training procedure is summarized in Algorithm~\ref{alg}.
We first train a ``teacher'' $\textsc{Filter}$ based on clean labels in the source language and the transferred ``noisy'' labels in the target language (if available) with loss from Eqn. (\ref{eqn:logit}).
This $\textsc{Filter}$ will then be used as a teacher to generate pseduo soft-labels to regularize a second $\textsc{Filter}$ (student) trained from scratch.
As the noise mainly comes from translated text, we only add soft labels in the target language during the training of the second $\textsc{Filter}$. Specifically,
\begin{eqnarray}
    \nonumber
    \boldsymbol{p}^s_{tea}, \boldsymbol{p}^t_{tea} &=& \textsc{Filter}_{tea}\, (\mathbf{S}, \mathbf{T}), \\
    \nonumber
    \boldsymbol{p}^s_{stu}, \boldsymbol{p}^t_{stu} &=& \textsc{Filter}_{stu}\, (\mathbf{S}, \mathbf{T}), \\
    \mathcal{L}^{kl} &=& \text{Loss}_{KL} (\boldsymbol{p}^t_{tea}, \boldsymbol{p}^t_{stu} ), 
\end{eqnarray}
where $\text{Loss}_{KL}$ denotes KL divergence. The soft label $\boldsymbol{p}^t_{tea}$ is fixed when training the student $\textsc{Filter}$, which is used for final prediction. When no labels can be transferred to the target language, this method helps the model receive more gradients on the target language, instead of purely on the source side, thus reducing the domain gap between languages. When labels can be transferred, it serves as a smoothing or regularization term appended to the supervised losses.
By merging the self-teaching loss, our final training objective for the student $\textsc{Filter}$ is summarized as:
\begin{equation}
    \mathcal{L}^{final}=\mathcal{L}^s+\lambda \mathcal{L}^t + (1-\lambda) \mathcal{L}^{kl},
\end{equation}
where $\lambda$ is a hyper-parameter to tune, and $\lambda$ is set to zero when no labels in the target languages can be transferred from the source language (\emph{e.g.}, for NER and POS tagging).

\subsection{Inference}
During inference, we pair the text input in the target language with the translated text in the source language, so that $\textsc{Filter}$ can fuse the information from both languages.
For classification tasks, we use the probabilities from either source or target language for prediction. However, for structured prediction and question answering tasks, only the probabilities from the target language can be used for prediction, as the tagging order is different between languages, and the answers are also difficult to evaluate if in different languages. 
Therefore, for simplicity, we consistently use the probabilities $\boldsymbol{p}^t_{stu}$ from the target language for final prediction.

%% file: exp.tex
\input{tables/dataset_statistics}
\input{tables/xtreme_leaderboard}
\input{tables/xglue_leaderboard}
\section{Experiments}
In this section, we present experimental results on the XTREME and XGLUE benchmarks and provide detailed analysis on the effectiveness of \textsc{Filter}.

\subsection{Datasets}
There are nine datasets in both XTREME~\cite{hu2020xtreme} and XGLUE~\cite{liang2020xglue} benchmarks for cross-lingual language understanding, which can be grouped into four categories (Classification, Structured Prediction, QA, and Retrieval). The statistics of each dataset is summarized in Table~\ref{tbl:stat}. Note that cross-lingual language generation tasks in XGLUE are not included. 

\noindent\textbf{Cross-lingual Sentence Classification} includes two common tasks: ($i$) Cross-lingual Natural Language Inference (XNLI)~\cite{conneau2018xnli}, and ($ii$) Cross-lingual Paraphrase Adversaries from Word Scrambling (PAWS-X)~\cite{yang2019paws}. 
In XGLUE, they further include another four practical tasks selected from Search, Ads and News scenarios: News Classification, Query-Ad Matching, Web Page Ranking and QA Matching.

\noindent\textbf{Cross-lingual Structured Prediction} includes two tasks: POS tagging and NER. 
In XTREME, the Wikiann dataset~\cite{pan2017cross} is used for experiments, and in XGLUE, they use a subset of two tasks from CoNLL-2002~\cite{tjong-kim-sang-2002-introduction} and CoNLL-2003 NER~\cite{tjong-kim-sang-de-meulder-2003-introduction}.

\noindent\textbf{Cross-lingual Question Answering} includes three tasks: ($i$) Cross-lingual Question Answering (XQuAD)~\cite{Artetxe:etal:2019}, ($ii$) Multilingual Question Answering (MLQA)~\cite{lewis2019mlqa}, and ($iii$) the gold passage version of the Typologically Diverse Question Answering dataset (TyDiQA-GoldP)~\cite{clark2020tydi}.

\noindent\textbf{Cross-lingual Sentence Retrieval} includes two tasks: BUCC~\cite{zweigenbaum2018overview} and Tatoeba~\cite{artetxe2019massively}. 
For leaderboard submission, we apply models trained on XNLI directly on these two datasets for inference.

\input{tables/full_results}

\subsection{Implementation Details}

Our implementation is based on HuggingFace's Transformers~\cite{Wolf2019HuggingFacesTS}.
We leverage the pre-trained XLM-R model~\cite{conneau2019unsupervised} to initialize our \textsc{Filter}, which contains 24 layers, each layer with 1,024 hidden states. For fair comparison to XLM-R, each transformer layer in \textsc{Filter} is shared for encoding both source and target languages, so that the total number of parameters are exactly the same as XLM-R.

We conduct experiments on 8 Nvidia V100-32GB GPU cards for model finetuning, and set batch size to 64 for all tasks. 
For self-teaching loss, we set the weight of the KL loss to 1.0 for structured prediction tasks where no labels are available in the target language.
We set the weight of KL loss for classification and QA tasks to 0.5 and 0.1 respectively, by searching over [0.1, 0.3, 0.5]. 
As the official XTREME repo\footnote{https://github.com/google-research/xtreme} does not provide translated target language data for POS and NER,
we use Microsoft Machine Translator\footnote{https://azure.microsoft.com/en-us/services/cognitive-services/translator/} for translation.
More details on translation data and model hyper-parameters are provided in Appendix.


\input{tables/xnli_results}


\subsection{Baselines}
We compare \textsc{Filter} with previous state-of-the-art multilingual models: 
\begin{itemize}[itemsep=1pt,topsep=2pt,leftmargin=12pt]
\item \textbf{Pre-trained models}: \textit{mBERT}~\cite{devlin-etal-2019-bert}, \textit{XLM}~\cite{lample2019cross}, \textit{XLM-R}~\cite{conneau2019unsupervised}, \textit{MMTE}~\cite{siddhant2020evaluating}, \textit{InfoXLM}~\cite{chi2020infoxlm},
\textit{Unicoder}~\cite{huang2019unicoder} pre-train Transformer models on large-scale multi-lingual dataset including machine translation data.
\item \textbf{Data augmentation}: 	X-STILTs~\cite{phang2020english} first finetunes XLM-R on an additional intermediate auxiliary task, then further finetunes on the target task.
\item \textbf{Translate-train}~\cite{hu2020xtreme} finetunes cross-lingual pre-trained language model XLM-R on  English training data and all translated data by using Google's in-house Machine Translation system.

\end{itemize}

\begin{figure*}[t]
\centering
\includegraphics[width=\linewidth]{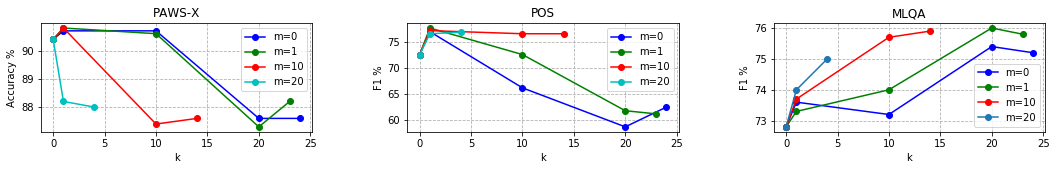}
\caption{Results on the dev set of PAWS-X, POS and MLQA  with different $m$ and $k$ values.}
\label{fig:abl_km}
\end{figure*}

\input{tables/transfer_gap}

\subsection{Experimental Results}
Table~\ref{tbl:leaderboard} and \ref{tbl:xglue_leaderboard} summarizes our results on XTREME and XGLUE, outperforming all the leaderboard submissions.
On XTREME, compared to the unpublished state-of-the-art VECO approach, \textsc{Filter} outperforms by 2.8/1.5/1.3/4.0 points on the four categories respectively, achieving an average score of 77.0, an absolute 2.2-point improvement.
Compared to the XLM-R baseline, we achieve an absolute 8.8-point improvement (77.0 vs. 68.2), which is a significant margin. On XGLUE, compared to the Unicoder baseline, \textsc{Filter} achieves an absolute 4.0-point improvement.

Table~\ref{tbl:detailed_results_full} provides more detailed results on different tasks in XTREME.
First, we build a strong translate-train baseline using XLM-R as the backbone, which already outperforms previous state-of-the-art models by a significant margin on every dataset.
Second, compared to the translate-train XLM-R baseline, \textsc{Filter} further provides 0.9 and 2.28 points improvement on average on classification and question answering tasks.
Lastly, the self-teaching loss further boosts the performance of \textsc{Filter} on every dataset, especially on POS and NER tasks.

To provide a deeper look into the model performance across languages, Table~\ref{tbl:xnli_detailed_results} provides results on each language, taking the XNLI dataset as an example. Results show that \textsc{Filter} outperforms all baselines on each language.
Complete results on other datasets are provided in Appendix.

\subsection{Ablation Analysis}
Below, we provide a detailed analysis to better understand the effectiveness of \textsc{Filter} and the self-teaching loss on different tasks. 
In general, we observe that different tasks need different numbers of ``local'' transformer layers ($m$) and
intermediate fusion layers ($k$)
Furthermore, the self-teaching loss is helpful on all tasks, especially on tasks lacking labels in the target languages.

\noindent\textbf{Effect of Fusing Languages} 
As shown in Table~\ref{tbl:detailed_results_full} and discussed above, \textsc{Filter} outperforms the translate-train baseline by a significant margin on classification and QA datasets, demonstrating the effectiveness of fusing languages. 
For POS and NER, there is no translate-train baseline as labels are unavailable in translated target language. 
Nonetheless, \textsc{Filter} improves XLM-R by 2.9 and 1.3 points, thanks to the use of intermediate cross-attention between language pair.
For the simple concatenation baseline, its performance can be analyzed by setting $m=0, k=24$ in Figure~\ref{fig:abl_km}. Compared to \textsc{Filter}, the performance drops 2.5/15.2 points on PAWS-X and POS datasets.
For MLQA, there is only a minor drop.
We hypothesize that for simple classification tasks, single-language input already provides rich information, while concatenating the paired language input directly at the very beginning introduces more noise, therefore making the model more difficult to train.
Overall, performing cross-attention between the language pair in intermediate layers performs the best. 

\noindent\textbf{Effect of Intermediate Fusion Layers}
Figure~\ref{fig:abl_km} shows the results on the dev sets with different $k$ and $m$ combinations (see Figure~\ref{fig:model} for its definition). 
We perform experiments on PAWS-X, POS and MLQA, and consider them as representative datasets for classification, structured prediction and question answering tasks.
For MLQA, performance is consistently improved with the number of intermediate fusion layers increasing, resulting in 2.6 points improvement from $k=1$ to $k=20$ when $m$ is set to 1.
By contrast, the performance on PAWS-X and POS drops significantly when the number of intermediate fusion layers increases.
For example, when $m$ is set to 1, accuracy decreases by 2.5/16.5 points from $k=1$ to $k=24$ on PAWS-X and POS datasets.

\noindent\textbf{Effect of Local Transformer Layers}
As shown in Figure~\ref{fig:abl_km}, for POS and MLQA, \textsc{Filter} performs better when using more local transformer layers.
For example, when $k$ is set to 10, we observe performance  improvement by setting $m$ to $0,1,10$ sequentially.
On the contrary, for PAWS-X, when $k=10$, the performance of setting $m = 0, 1$ is better than setting $m=10$.
This suggests that we should use more local layers for complex tasks such as QA and structured prediction, and fewer local layers for classification tasks. 

\noindent\textbf{Effect of Self-teaching Loss}
As can be seen from Table~\ref{tbl:detailed_results_full}, 
for POS and NER, the use of self-teaching loss improves \textsc{Filter} by 0.7 and 1.0 points. This confirms that self-teaching loss is very helpful in addressing the no-label issue for target languages.
For classification and question answering tasks, we observe minor improvement, which is expected, as ground-truth labels are available for target languages, and adding the self-teaching loss only provides some label smoothing effect. 

\noindent\textbf{Cross-lingual Transfer Gap}
Table~\ref{tbl:trans_gap} shows analysis results of cross-lingual gap of different models, by calculating the difference between the performance on English test set and the average performance of other target languages.
We observe that \textsc{Filter} reduces the cross-lingual gap significantly among all tasks compared to mBERT, XLM-R and translate-train baselines.
The transfer learning gap of \textsc{Filter} is reduced by additional 2.5 and 10.6 points on average for classification and QA tasks, respectively, compared to the translate-train baseline respectively. 
For structured prediction tasks, the gap reduces even further, but a large gap still exists, indicating that this task demands stronger cross-lingual transfer. 

\section{Conclusion}
We present \textsc{Filter}, a new approach for cross-lingual language understanding that  
first encodes paired language input independently, then fuses them in the intermediate layers of XLM, and finally performs further language-specific encoding.
An additional self-teaching loss is proposed for enhanced model training. By combining \textsc{Filter} and self-teaching loss, we achieve new state of the art on the challenging XTREME and XGLUE benchmarks.
Future work points to more effective ways of automatically discovering the best configuration of \textsc{Filter} for different cross-lingual tasks.

%% file: tables/dataset_statistics.tex
\begin{table}[t!]
\resizebox{1.0\columnwidth}{!}{

\begin{tabular}{lcccccc}
\toprule
Benchmark & Task & Dataset  & \#train & \#dev & \#test & \#languages \\
\midrule
\multirow{9}{*}{XTREME} &
\multirow{2}{*}{Classification} & XNLI & 392K & 2.5K & 5K & 15 \\
& & PAWS-X & 49.4K & 2K & 2K & 7 \\
\cmidrule{2-7}
& \multirow{2}{*}{Struct. pred. } & POS & 21K & 4K & 47-20K & 33 \\
& & NER & 20K & 10K & 1K-10K & 40 \\
\cmidrule{2-7}
& \multirow{3}{*}{QA} & XQuAD & \multirow{2}{*}{87K} &  \multirow{2}{*}{34K} & 1190 & 11 \\
& & MLQA & & & 4.5K-11K & 7 \\
& & TyDiQA-GoldP &  3.7K & 0.6K & 0.3K–2.7K & 9 \\
\cmidrule{2-7}
& \multirow{2}{*}{Retrieval} & BUCC & - & - & 1.9K–14K & 5 \\
& & Tatoeba & - & - & 1K & 33 \\
\midrule
\multirow{9}{*}{XGLUE} & 
\multirow{6}{*}{Classification} & XNLI & 392K & 2.5K & 5K & 15 \\
& & PAWS-X & 49.4K & 2K & 2K & 4 \\
& & NC & 100K & 10K & 10K & 5 \\
& & QADSM & 100K & 10K & 10K & 3 \\
& & WPR & 100K & 10K & 10K & 7 \\
& & QAM & 100K & 10K & 10K & 3 \\
\cmidrule{2-7}
& \multirow{2}{*}{Struct. pred. } & POS & 25.4K & 1.0K & 0.9K & 18 \\
& & NER & 15.0K & 2.8K & 3.4K & 4 \\
\cmidrule{2-7}
& QA & MLQA & 87.6K & 0.6K & 5.7K & 7 \\
\bottomrule
\end{tabular}}
\caption{Statistics of the datasets in XTREME and XGLUE. \#train, \#dev and \#test are the numbers of examples in the training, dev and test sets, respectively. For dev and test set, the number is for each target language. \#languages is the number of target languages in the test set. Note that language generation tasks in XGLUE are not included.}
\label{tbl:stat}
\end{table}

%% file: tables/xtreme_leaderboard.tex
\begin{table*}[!htb]
\centering
\begin{adjustbox}{scale=0.95,center}
\begin{tabular}{lccccc}
\toprule
Model & Avg & Pair sentence & Structured prediction & Question answering & Sentence retrieval \\
\midrule
XLM             & 55.8 & 75.0 & 65.6  & 43.9 & 44.7 \\
MMTE            & 59.3 & 74.3 & 65.3  & 52.3 & 48.9 \\
mBERT           & 59.6 & 73.7 & 66.3  & 53.8 & 47.7 \\
XLM-R           & 68.2 & 82.8  & 69.0 & 62.3 & 61.6 \\
X-STILTs         & 73.5 & 83.9  & 69.4 & 67.2 & 76.5 \\
VECO$^\dagger$            & 74.8	& 84.7 & 70.4 & 67.2 & 80.5 \\
\midrule
\textsc{Filter}            & \textbf{77.0} & \textbf{87.5} & \textbf{71.9} & \textbf{68.5} & \textbf{84.5} \\
\bottomrule
\end{tabular}
\end{adjustbox}
\caption{Results on the test set of XTREME. \textsc{Filter} achieves new state of the art at the time of submission (Sep. 8, 2020). For TydiQA-GoldP dataset, we use additional SQuAD v1.1 English training data. The score on question answering is calculated by the average of EM and F1 scores on three datasets. ($\dagger$) indicates unpublished work.
}
\label{tbl:leaderboard}
\end{table*}

%% file: tables/xglue_leaderboard.tex
\begin{table*}[t!]
\centering
\begin{adjustbox}{scale=0.95,center}
\begin{tabular}{lcccccccccc}
\toprule

Model & Avg & NER & POS	& NC & MLQA & XNLI & PAWS-X & QADSM & WPR & QAM \\
\midrule
Unicoder & 76.1 & 79.7 & 79.6	& \textbf{83.5} & 66.0 & 75.3 & 90.1 & 68.4 & 73.9 & 68.9 \\
\midrule
\textsc{Filter} & \textbf{80.1} & \textbf{82.6} & \textbf{81.6} & \textbf{83.5} & \textbf{76.2} & \textbf{83.9} & \textbf{93.8} & \textbf{71.4} & \textbf{74.7} & \textbf{73.4} \\
\bottomrule
\end{tabular}
\end{adjustbox}
\caption{Results on the test set of XGLUE. \textsc{Filter} achieves new state of the art at the time of submission (Sep. 14, 2020). Note that cross-lingual language generation tasks are not included.
Leaderboard: \href{https://microsoft.github.io/XGLUE}{https://microsoft.github.io/XGLUE}.}
\label{tbl:xglue_leaderboard}
\end{table*}

%% file: tables/full_results.tex
\begin{table*}[t!]
\centering
\begin{adjustbox}{scale=0.95,center}
\begin{tabular}{lccccccc}
\toprule
\multirow{2}{*}{Model} & \multicolumn{2}{c}{Pair sentence} & \multicolumn{2}{c}{Structured prediction} & \multicolumn{3}{c}{Question answering}  \\
& XNLI & PAWS-X & POS & NER & XQuAD & MLQA & TyDiQA-GoldP \\
\midrule
Metrics & Acc. & Acc. & F1 & F1 & F1 / EM & F1 / EM & F1 / EM  \\
\midrule
\multicolumn{8}{l}{\emph{Cross-lingual zero-shot transfer (models are trained on English data)}} \\
\midrule
mBERT          & 65.4 & 81.9 & 70.3 & 62.2 & 64.5 / 49.4 & 61.4 / 44.2 & 59.7 / 43.9 \\
XLM            & 69.1 & 80.9 & 70.1 & 61.2 & 59.8 / 44.3 & 48.5 / 32.6 & 43.6 / 29.1 \\
XLM-R          & 79.2 & 86.4 & 72.6 & 65.4 & 76.6 / 60.8 & 71.6 / 53.2 & 65.1 / 45.0 \\
InfoXLM        & 81.4 & - & - & - & - / - & 73.6 / 55.2 & - / - \\
X-STILTs        & 80.0 & 87.9 & 74.4 & 64.0 & 78.7 / 63.3 & 72.4 / 53.7 & \textbf{76.0} / \textbf{59.5} \\
\midrule
\multicolumn{8}{l}{\emph{Translate-train (models are trained on English training data and its translated data on the target language)}} \\
\midrule
mBERT                    & 74.0 & 86.3 & -    & -    & 70.0 / 56.0 & 65.6 / 48.0 & 55.1 / 42.1 \\
mBERT, multi-task        & 75.1 & 88.9 & -    & -    & 72.4 / 58.3 & 67.6 / 49.8 & 64.2 / 49.3 \\
XLM-R, multi-task (Ours)  & 82.6 & 90.4 &  -   & -    & 80.2 / 65.9 & 72.8 / 54.3 & 66.5 / 47.7 \\
\midrule
\textsc{Filter} (Ours)                & 83.6   & 91.2 & 75.5 & 66.7 & 82.3 / 67.8  & 75.8 / 57.2 & 68.1 / 49.7 \\
\textsc{Filter} + Self-Teaching (Ours)  & \textbf{83.9}   & \textbf{91.4} & \textbf{76.2} & \textbf{67.7} & \textbf{82.4} / \textbf{68.0}  & \textbf{76.2} / \textbf{57.7} & 68.3 / 50.9 \\
\bottomrule
\end{tabular}
\end{adjustbox}
\caption{Overall test results on three different categories of cross-lingual language understanding tasks. Results of mBERT~\cite{devlin-etal-2019-bert}, XLM~\cite{lample2019cross} and XLM-R~\cite{conneau2019unsupervised} are from XTREME~\cite{hu2020xtreme}. InfoXLM~\cite{chi2020infoxlm} only provides results on XNLI and MLQA. We also experimented on translate-train with XLM-R as an additional baseline for fair comparison with \textsc{Filter}. 
}
\label{tbl:detailed_results_full}
\end{table*}

%% file: tables/xnli_results.tex
\begin{table*}[t!]
\centering
\resizebox{\linewidth}{!}{
\begin{tabular}{l|ccccccccccccccc|c}
\toprule
Model & en & ar & bg & de & el & es & fr & hi & ru & sw & th & tr & ur & vi & zh & avg \\
\midrule
mBERT                     & 80.8 & 64.3 & 68.0 & 70.0 & 65.3 & 73.5 & 73.4 & 58.9 & 67.8 & 49.7 & 54.1 & 60.9 & 57.2 & 69.3 & 67.8 & 65.4 \\
MMTE                      & 79.6 & 64.9 & 70.4 & 68.2 & 67.3 & 71.6 & 69.5 & 63.5 & 66.2 & 61.9 & 66.2 & 63.6 & 60.0 & 69.7 & 69.2 & 67.5 \\
XLM                       & 82.8 & 66.0 & 71.9 & 72.7 & 70.4 & 75.5 & 74.3 & 62.5 & 69.9 & 58.1 & 65.5 & 66.4 & 59.8 & 70.7 & 70.2 & 69.1 \\
XLM-R                     & 88.7 & 77.2 & 83.0 & 82.5 & 80.8 & 83.7 & 82.2 & 75.6 & 79.1 & 71.2 & 77.4 & 78.0 & 71.7 & 79.3 & 78.2 & 79.2 \\
\midrule
\shortstack{XLM-R (\emph{translate-train})}  & 88.6 & 82.2 & 85.2 & 84.5 & 84.5 & 85.7 & 84.2 & 80.8 & 81.8 & 77.0 & 80.2 & 82.1 & 77.7 & 82.6 & 82.7 & 82.6 \\
\textsc{Filter}          & \textbf{89.7} & 83.2 & 86.2 & 85.5 & 85.1 & \textbf{86.6} & 85.6 & 80.9 & 83.4 & 78.2 & \textbf{82.2} & 83.1 & 77.4 & 83.7 & 83.7 & 83.6 \\
\textsc{Filter} + Self-Teaching & 89.5 & \textbf{83.6} & \textbf{86.4} & \textbf{85.6} & \textbf{85.4} & \textbf{86.6} & \textbf{85.7} & \textbf{81.1} & \textbf{83.7} & \textbf{78.7} & 81.7 & \textbf{83.2} & \textbf{79.1} & \textbf{83.9} & \textbf{83.8} & \textbf{83.9} \\

\bottomrule
\end{tabular}
}
\caption{XNLI accuracy scores for each language. Results of mBERT, MMTE, XLM and XLM-R are from XTREME~\cite{hu2020xtreme}. \emph{mtl} denotes translate-train in multi-task version.}
\label{tbl:xnli_detailed_results}
\end{table*}

%% file: tables/transfer_gap.tex
\begin{table*}[t!]
\centering
\resizebox{1.55\columnwidth}{!}{
\begin{tabular}{lccccc|c|cc}
\toprule
Model         & XNLI & PAWS-X & XQuAD & MLQA & TyDiQA-GoldP & Avg  & POS  & NER \\
\midrule
mBERT\cite{hu2020xtreme}           & 16.5 & 14.1   & 25.0  & 27.5 & 22.2 & 21.1 & 25.5 & 23.6 \\

XLM-R\cite{hu2020xtreme}            & 10.2 & 12.4   & 16.3 & 19.1 & 13.3 & 14.3 & 24.3 & 19.8 \\
Translate-train\cite{hu2020xtreme}  & 7.3  & 9.0    & 17.6 & 22.2 & 24.2 & 16.1 & - & - \\
\midrule
\textsc{Filter}         & \textbf{6.0}    & \textbf{5.2}    & \textbf{7.3}  & \textbf{15.7} & \textbf{9.2} & \textbf{8.7} & \textbf{19.7} & \textbf{16.3} \\
\bottomrule
\end{tabular}
}
\caption{Analysis on cross-lingual transfer gap of different models on XTREME benchmark (except for retrieval  task). A lower gap indicates a better cross-lingual transfer model. 
The average score (Avg) is calculated on all classification and QA tasks. 
}
\label{tbl:trans_gap}
\end{table*}

%% file: appendix.tex
\appendix

\section{Hyper-parameters}
\label{sec:hyper-p}
For XNLI, PAWS-X and TyDiQA-Gold, we finetune 4 epochs. For MLQA and XQuAD, we finetune 2 epochs. 
To select the best $k$ and $m$ for each dataset, we choose PAWS-X, POS and MLQA as the representative datasets for each category. Then, we perform grid search over $k$ and $m$ from [1, 10, 20, 24] and [0, 1, 10, 20] on the dev set, respectively, and apply the best hyper-parameters for all tasks in each category. Note that we keep $k + m \le 24$. 
After choosing the best $k$ and $m$, learning rate is the only hyper-parameter tuned for \textsc{Filter}. We select the model with the best average result over all the languages on the dev sets, by searching the learning rate over [3e-6, 5e-6, 1e-5]. 
We use the hyper-parameters learned from MLQA for XQuAD test set, which does not have a dev set.

\section{Translation Data}
During training, we use the provided English training data as the source language. The translated target-language training data of XNLI and PAWS-X are provided in the original datasets.
For POS and NER, we use Microsoft Machine Translator\footnote{https://azure.microsoft.com/en-us/services/cognitive-services/translator/} to translate English training data to target languages. As the translator does not cover all target languages, we exclude paired training data in the following languages: Basque, Javanese, Georgian, Burmese, Tagalog and Yoruba.
For XQuAD, MLQA and TyDiQA-GoldP, we use the translation data provided by the official XTREME repo\footnote{https://github.com/google-research/xtreme}.
For leaderboard submission, we use additional SQuAD v1.1 training data during finetuning on TyDiQA-Gold, as the original training set only contains 3K training samples.
During inference, we automatically translate the target-language test data to English using the aforementioned translator. 
For POS and NER, we use the original target-language text itself if the target languages are not covered by the translator.

\section{Results for Each Dataset and Language}
Below, we provide detailed results for each dataset and language. Results of mBERT, XLM, MMTE and XLM-R are from XTREME~\cite{hu2020xtreme}.

\begin{table}[h!]
\centering
\resizebox{\columnwidth}{!}{
\begin{tabular}{l|ccccccc|c}
\toprule
Model & en & de & es & fr & ja & ko & zh & avg \\
\midrule
mBERT & 94.0& 85.7& 87.4& 87.0& 73.0& 69.6& 77.0 & 81.9 \\
XLM & 94.0& 85.9& 88.3& 87.4& 69.3& 64.8& 76.5 & 80.9 \\
MMTE & 93.1& 85.1& 87.2& 86.9& 72.0& 69.2& 75.9 & 81.3 \\
XLM-R & 94.7& 89.7& 90.1& 90.4& 78.7& 79.0& 82.3 & 86.4 \\
\midrule
\textsc{Filter} & 96.5 & 92.5 & 93.0 & 93.8 & 86.7 & 87.1 & 88.3 & 91.2 \\
\textsc{Filter} + Self-Teaching & 95.9 & 92.8 & 93.0 & 93.7 & 87.4 & 87.6 & 89.6 & 91.5 \\
\bottomrule
\end{tabular}
}
\caption{PAWS-X accuracy scores for each language.}
\label{tbl:xnli}
\end{table}

\begin{table*}[]
\centering
\resizebox{\linewidth}{!}{
\begin{tabular}{l|ccccccccccc|c}
\toprule
Model & en & ar & de & el & es & hi & ru & th & tr & vi & zh & avg \\
\midrule
mBERT & 83.5 / 72.2 &  61.5 / 45.1 &  70.6 / 54.0 &  62.6 / 44.9 &  75.5 / 56.9 &  59.2 / 46.0 &  71.3 / 53.3 &  42.7 / 33.5 &  55.4 / 40.1 &  69.5 / 49.6 &  58.0 / 48.3 &  64.5 / 49.4 \\
XLM & 74.2 / 62.1 &  61.4 / 44.7 &  66.0 / 49.7 &  57.5 / 39.1 &  68.2 / 49.8 &  56.6 / 40.3 &  65.3 / 48.2 &  35.4 / 24.5 &  57.9 / 41.2 &  65.8 / 47.6 &  49.7 / 39.7 &  59.8 / 44.3 \\
MMTE & 80.1 / 68.1 &  63.2 / 46.2 &  68.8 / 50.3 &  61.3 / 35.9 &  72.4 / 52.5 &  61.3 / 47.2 &  68.4 / 45.2 &  48.4 / 35.9 &  58.1 / 40.9 &  70.9 / 50.1 &  55.8 / 36.4 &  64.4 / 46.2 \\
XLM-R & 86.5 / 75.7 &  68.6 / 49.0 &  80.4 / 63.4 &  79.8 / 61.7 &  82.0 / 63.9 &  76.7 / 59.7 &  80.1 / 64.3 &  74.2 / 62.8 &  75.9 / 59.3 &  79.1 / 59.0 &  59.3 / 50.0 &  76.6 / 60.8 \\
\midrule
\textsc{Filter} & 85.6 / 73.0 & 79.8 / 61.3 & 82.5 / 66.2 & 82.6 / 64.6 & 84.8 / 67.4 & 83.1 / 66.5 & 82.5 / 66.8 & 80.7 / 73.9 & 81.2 / 65.7 & 83.3 / 64.1 & 78.9 / 75.7 & 82.3 / 67.8 \\
\textsc{Filter} + Self-Teaching & 86.4 / 74.6 & 79.5 / 60.7 & 83.2 / 67.0 & 83.0 / 64.6 & 85.0 / 67.9 & 83.1 / 66.6 & 82.8 / 67.4 & 79.6 / 73.2 & 80.4 / 64.4 & 83.8 / 64.7 & 79.9 / 77.0 & 82.4 / 68.0 \\
\bottomrule
\end{tabular}
}
\caption{XQuAD results (F1 / EM) for each language.}
\label{tbl:xnli}
\end{table*}

\begin{table*}[]
\centering
\resizebox{\linewidth}{!}{
\begin{tabular}{l|ccccccc|c}
\toprule
Model & en & ar & de & es & hi & vi & zh & avg \\
\midrule
mBERT & 80.2 / 67.0 &  52.3 / 34.6 &  59.0 / 43.8 &  67.4 / 49.2 &  50.2 / 35.3 &  61.2 / 40.7 &  59.6 / 38.6 & 61.4 / 44.2\\
XLM & 68.6 / 55.2 &  42.5 / 25.2 &  50.8 / 37.2 &  54.7 / 37.9 &  34.4 / 21.1 &  48.3 / 30.2 &  40.5 / 21.9 & 48.5 / 32.6\\
MMTE & 78.5 / – &  56.1 / – &  58.4 / – &  64.9 / – &  46.2 / – &  59.4 / – &  58.3 / –  & 60.3 / 41.4\\
XLM-R & 83.5 / 70.6 &  66.6 / 47.1 &  70.1 / 54.9 &  74.1 / 56.6 &  70.6 / 53.1 &  74.0 / 52.9 &  62.1 / 37.0 & 71.6 / 53.2\\
\midrule
\textsc{Filter} & 83.5 / 70.3 & 71.8 / 51.0 & 74.6 / 59.8 & 77.9 / 60.2 & 76.1 / 57.7 & 77.7 / 57.2 & 69.0 / 44.2 & 75.8 / 57.2 \\
\textsc{Filter} + Self-Teaching & 84.0 / 70.8 & 72.1 / 51.1 & 74.8 /60.0 & 78.1 / 60.1 & 76.0 / 57.6 & 78.1 /57.5 & 70.5 / 47.0 & 76.2 / 57.7\\
\bottomrule
\end{tabular}
}
\caption{MLQA results (F1 / EM) for each language.}
\label{tbl:xnli}
\end{table*}

\begin{table*}[]
\centering
\resizebox{\linewidth}{!}{
\begin{tabular}{l|ccccccccc|c}
\toprule
Model &  en & ar & bn & fi & id & ko & ru & sw & te & avg \\
\midrule
mBERT & 75.3 / 63.6 &  62.2 / 42.8 &  49.3 / 32.7 &  59.7 / 45.3 &  64.8 / 45.8 &  58.8 / 50.0 &  60.0 / 38.8 &  57.5 / 37.9 &  49.6 / 38.4 &  59.7 / 43.9 \\
XLM & 66.9 / 53.9 &  59.4 / 41.2 &  27.2 / 15.0 &  58.2 / 41.4 &  62.5 / 45.8 &  14.2 / 5.1 &  49.2 / 30.7 &  39.4 / 21.6 &  15.5 / 6.9 &  43.6 / 29.1 \\
MMTE & 62.9 / 49.8 &  63.1 / 39.2 &  55.8 / 41.9 &  53.9 / 42.1 &  60.9 / 47.6 &  49.9 / 42.6 &  58.9 / 37.9 &  63.1 / 47.2 &  54.2 / 45.8 &  58.1 / 43.8 \\
XLM-R & 71.5 / 56.8 &  67.6 / 40.4 &  64.0 / 47.8 &  70.5 / 53.2 &  77.4 / 61.9 &  31.9 / 10.9 &  67.0 / 42.1 &  66.1 / 48.1 &  70.1 / 43.6 &  65.1 / 45.0 \\
\midrule
\textsc{Filter} & 71.9 / 58.9 & 73.7 / 47.9 & 68.7 / 53.1 & 71.2 / 54.9 & 77.9 / 59.8 & 33.0 / 12.3 & 68.7 / 45.9 & 78.7 / 66.1 & 69.4 / 48.6 & 68.1 / 49.7 \\
\textsc{Filter} + Self-Teaching & 72.4 / 59.1 & 72.8 / 50.8 & 70.5 / 56.6 & 73.3 / 57.2 & 76.8 / 59.8 & 33.1 / 12.3 & 68.9 / 46.6 & 77.4 / 65.7 & 69.9 / 50.4 & 68.3 / 50.9 \\
\bottomrule
\end{tabular}
}
\caption{TyDiQA-GolP results (F1 / EM) for each language.}
\label{tbl:xnli}
\end{table*}

\begin{table*}[]
\centering
\resizebox{\linewidth}{!}{
\begin{tabular}{l|ccccccccccccccccc}
\toprule
Model &  af & ar & bg & de & el & en & es & et & eu & fa &  fi & fr & he & hi & hu & id & it \\
\midrule
mBERT & 86.6& 56.2& 85.0& 85.2& 81.1& 95.5& 86.9& 79.1& 60.7& 66.7& 78.9& 84.2& 56.2& 67.2& 78.3& 71.0& 88.4 \\
XLM & 88.5& 63.1& 85.0& 85.8& 84.3& 95.4& 85.8& 78.3& 62.8& 64.7& 78.4& 82.8& 65.9& 66.2& 77.3& 70.2& 87.4 \\
MMTE & 86.2& 65.9& 87.2& 85.8& 77.7& 96.6& 85.8& 81.6& 61.9& 67.3& 81.1& 84.3& 57.3& 76.4& 78.1& 73.5& 89.2 \\
XLM-R & 89.8& 67.5& 88.1& 88.5& 86.3& 96.1& 88.3& 86.5& 72.5& 70.6& 85.8& 87.2& 68.3& 76.4& 82.6& 72.4& 89.4 \\
\midrule
\textsc{Filter} & 88.5 & 66.0 & 87.6 & 89.0 & 88.1 & 96.0 & 89.0 & 85.9 & 76.8 & 70.7 & 85.9 & 87.8 & 64.9 & 75.4 & 82.5 & 72.6 & 88.6 \\
\textsc{Filter} + Self-Teaching & 88.7 & 66.1 & 88.5 & 89.2 & 88.3 & 96.0 & 89.1 & 86.3 & 78.0 & 70.8 & 86.1 & 88.9 & 64.9 & 76.7 & 82.6 & 72.6 & 89.8 \\
\midrule
& ja & kk & ko & mr & nl & pt & ru & ta & te & th & tl & tr & ur & vi & yo & zh & avg \\
\midrule
mBERT & 49.2 & 70.5 & 49.6 & 69.4 & 88.6 & 86.2 & 85.5 & 59.0 & 75.9 & 41.7 & 81.4 & 68.5 & 57.0 & 53.2 & 55.7 & 61.6 & 71.5 \\
XLM & 49.0 & 70.2 & 50.1 & 68.7 & 88.1 & 84.9 & 86.5 & 59.8 & 76.8 & 55.2 & 76.3 & 66.4 & 61.2 & 52.4 & 20.5 & 65.4 & 71.3 \\
MMTE & 48.6 & 70.5 & 59.3 & 74.4 & 83.2 & 86.1 & 88.1 & 63.7 & 81.9 & 43.1 & 80.3 & 71.8 & 61.1 & 56.2 & 51.9 & 68.1 & 73.5 \\
XLM-R & 15.9 & 78.1 & 53.9 & 80.8 & 89.5 & 87.6 & 89.5 & 65.2 & 86.6 & 47.2 & 92.2 & 76.3 & 70.3 & 56.8 & 24.6 & 25.7 & 73.8 \\
\midrule
\textsc{Filter} & 38.4 & 79.5 & 53.0 & 84.7 & 89.3 & 88.1 & 90.4 & 64.8 & 87.6 & 54.5 & 93.1 & 76.3 & 68.6 & 57.6 & 39.2 & 52.6 & 76.2 \\
\textsc{Filter} + Self-Teaching & 40.4 & 80.4 & 53.3 & 86.4 & 89.4 & 88.3 & 90.5 & 65.3 & 87.3 & 57.2 & 94.1 & 77.0 & 70.9 & 58.0 & 43.1 & 53.1 & 76.9 \\
\bottomrule
\end{tabular}
}
\caption{POS results (Accuracy) for each language.}
\label{tbl:xnli}
\end{table*}

\begin{table*}[]
\centering
\resizebox{\linewidth}{!}{
\begin{tabular}{l|cccccccccccccccccccc}
\toprule
Model & en & af & ar & bg & bn & de & el & es & et & eu & fa & fi & fr & he & hi & hu & id & it & ja & jv \\
\midrule
mBERT & 85.2 & 77.4 & 41.1 & 77.0 & 70.0 & 78.0 & 72.5 & 77.4 & 75.4 & 66.3 & 46.2 & 77.2 & 79.6 & 56.6 & 65.0 & 76.4 & 53.5 & 81.5 & 29.0 & 66.4 \\
XLM & 82.6 & 74.9 & 44.8 & 76.7 & 70.0 & 78.1 & 73.5 & 74.8 & 74.8 & 62.3 & 49.2 & 79.6 & 78.5 & 57.7 & 66.1 & 76.5 & 53.1 & 80.7 & 23.6 & 63.0 \\
MMTE & 77.9 & 74.9 & 41.8 & 75.1 & 64.9 & 71.9 & 68.3 & 71.8 & 74.9 & 62.6 & 45.6 & 75.2 & 73.9 & 54.2 & 66.2 & 73.8 & 47.9 & 74.1 & 31.2 & 63.9 \\
XLM-R & 84.7 & 78.9 & 53.0 & 81.4 & 78.8 & 78.8 & 79.5 & 79.6 & 79.1 & 60.9 & 61.9 & 79.2 & 80.5 & 56.8 & 73.0 & 79.8 & 53.0 & 81.3 & 23.2 & 62.5 \\
\midrule
\textsc{Filter} & 83.3 & 78.7 & 56.2 & 83.3 & 75.4 & 79.0 & 79.7 & 75.6 & 80.0 & 67.0 & 70.3 & 80.1 & 79.6 & 55.0 & 72.3 & 80.2 & 52.7 & 81.6 & 25.2 & 61.8 \\
\textsc{Filter} + self-teaching & 83.5 & 80.4 & 60.7 & 83.5 & 78.4 & 80.4 & 80.7 & 74.0 & 81.0 & 66.9 & 71.3 & 80.2 & 79.9 & 57.4 & 74.3 & 82.2 & 54.0 & 81.9 & 24.3 & 63.5 \\
\midrule
& ka & kk & ko & ml & mr & ms & my & nl & pt & ru & sw & ta & te & th & tl & tr & ur & vi & yo & zh \\
\midrule
mBERT & 64.6 & 45.8 & 59.6 & 52.3 & 58.2 & 72.7 & 45.2 & 81.8 & 80.8 & 64.0 & 67.5 & 50.7 & 48.5 & 3.6 & 71.7 & 71.8 & 36.9 & 71.8 & 44.9 & 42.7 \\
XLM & 67.7 & 57.2 & 26.3 & 59.4 & 62.4 & 69.6 & 47.6 & 81.2 & 77.9 & 63.5 & 68.4 & 53.6 & 49.6 & 0.3 & 78.6 & 71.0 & 43.0 & 70.1 & 26.5 & 32.4 \\
MMTE & 60.9 & 43.9 & 58.2 & 44.8 & 58.5 & 68.3 & 42.9 & 74.8 & 72.9 & 58.2 & 66.3 & 48.1 & 46.9 & 3.9 & 64.1 & 61.9 & 37.2 & 68.1 & 32.1 & 28.9 \\
XLMR & 71.6 & 56.2 & 60.0 & 67.8 & 68.1 & 57.1 & 54.3 & 84.0 & 81.9 & 69.1 & 70.5 & 59.5 & 55.8 & 1.3 & 73.2 & 76.1 & 56.4 & 79.4 & 33.6 & 33.1 \\
\midrule
\textsc{Filter} & 70.0 & 50.6 & 63.8 & 67.3 & 66.4 & 68.1 & 60.7 & 83.7 & 81.8 & 71.5 & 68.0 & 62.8 & 56.2 & 1.5 & 74.5 & 80.9 & 71.2 & 76.2 & 40.4 & 35.9 \\
\textsc{Filter} + Self-Teaching & 71.0 & 51.1 & 63.8 & 70.2 & 69.8 & 69.3 & 59.0 & 84.6 & 82.1 & 71.1 & 70.6 & 64.3 & 58.7 & 2.4 & 74.4 & 83.0 & 73.4 & 75.8 & 42.9 & 35.4\\
\bottomrule
\end{tabular}
}
\caption{NER results (F1) for each language.}
\label{tbl:xnli}
\end{table*}

%% file: main.bbl
\begin{thebibliography}{33}
\providecommand{\natexlab}[1]{#1}
\providecommand{\url}[1]{\texttt{#1}}
\providecommand{\urlprefix}{URL }
\expandafter\ifx\csname urlstyle\endcsname\relax
  \providecommand{\doi}[1]{doi:\discretionary{}{}{}#1}\else
  \providecommand{\doi}{doi:\discretionary{}{}{}\begingroup
  \urlstyle{rm}\Url}\fi

\bibitem[{Artetxe, Ruder, and Yogatama(2020)}]{Artetxe:etal:2019}
Artetxe, M.; Ruder, S.; and Yogatama, D. 2020.
\newblock On the cross-lingual transferability of monolingual representations.
\newblock In \emph{Association for Computational Linguistics}.

\bibitem[{Artetxe and Schwenk(2019)}]{artetxe2019massively}
Artetxe, M.; and Schwenk, H. 2019.
\newblock Massively multilingual sentence embeddings for zero-shot
  cross-lingual transfer and beyond.
\newblock \emph{Transactions of the Association for Computational Linguistics}
  .

\bibitem[{Cao, Liu, and Wan(2020)}]{cao2020jointly}
Cao, Y.; Liu, H.; and Wan, X. 2020.
\newblock Jointly Learning to Align and Summarize for Neural Cross-Lingual
  Summarization.
\newblock In \emph{Association for Computational Linguistics}.

\bibitem[{Cer et~al.(2017)Cer, Diab, Agirre, Lopez-Gazpio, and
  Specia}]{cer2017semeval}
Cer, D.; Diab, M.; Agirre, E.; Lopez-Gazpio, I.; and Specia, L. 2017.
\newblock Semeval-2017 task 1: Semantic textual similarity-multilingual and
  cross-lingual focused evaluation.
\newblock \emph{arXiv preprint arXiv:1708.00055} .

\bibitem[{Chi et~al.(2020)Chi, Dong, Wei, Yang, Singhal, Wang, Song, Mao,
  Huang, and Zhou}]{chi2020infoxlm}
Chi, Z.; Dong, L.; Wei, F.; Yang, N.; Singhal, S.; Wang, W.; Song, X.; Mao,
  X.-L.; Huang, H.; and Zhou, M. 2020.
\newblock InfoXLM: An Information-Theoretic Framework for Cross-Lingual
  Language Model Pre-Training.
\newblock \emph{arXiv preprint arXiv:2007.07834} .

\bibitem[{Clark et~al.(2020)Clark, Choi, Collins, Garrette, Kwiatkowski,
  Nikolaev, and Palomaki}]{clark2020tydi}
Clark, J.~H.; Choi, E.; Collins, M.; Garrette, D.; Kwiatkowski, T.; Nikolaev,
  V.; and Palomaki, J. 2020.
\newblock TyDi QA: A benchmark for information-seeking question answering in
  typologically diverse languages.
\newblock \emph{Transactions of the Association for Computational Linguistics}
  .

\bibitem[{Conneau et~al.(2020)Conneau, Khandelwal, Goyal, Chaudhary, Wenzek,
  Guzm{\'a}n, Grave, Ott, Zettlemoyer, and Stoyanov}]{conneau2019unsupervised}
Conneau, A.; Khandelwal, K.; Goyal, N.; Chaudhary, V.; Wenzek, G.; Guzm{\'a}n,
  F.; Grave, E.; Ott, M.; Zettlemoyer, L.; and Stoyanov, V. 2020.
\newblock Unsupervised cross-lingual representation learning at scale.
\newblock In \emph{Association for Computational Linguistics}.

\bibitem[{Conneau et~al.(2018)Conneau, Lample, Rinott, Williams, Bowman,
  Schwenk, and Stoyanov}]{conneau2018xnli}
Conneau, A.; Lample, G.; Rinott, R.; Williams, A.; Bowman, S.~R.; Schwenk, H.;
  and Stoyanov, V. 2018.
\newblock XNLI: Evaluating cross-lingual sentence representations.
\newblock In \emph{Empirical Methods in Natural Language Processing}.

\bibitem[{Cui et~al.(2019)Cui, Che, Liu, Qin, Wang, and Hu}]{cui2019cross}
Cui, Y.; Che, W.; Liu, T.; Qin, B.; Wang, S.; and Hu, G. 2019.
\newblock Cross-lingual machine reading comprehension.
\newblock In \emph{Empirical Methods in Natural Language Processing}.

\bibitem[{Devlin et~al.(2019)Devlin, Chang, Lee, and
  Toutanova}]{devlin-etal-2019-bert}
Devlin, J.; Chang, M.-W.; Lee, K.; and Toutanova, K. 2019.
\newblock {BERT}: Pre-training of Deep Bidirectional Transformers for Language
  Understanding.
\newblock In \emph{North {A}merican Chapter of the Association for
  Computational Linguistics}.

\bibitem[{Faruqui and Dyer(2014)}]{faruqui2014improving}
Faruqui, M.; and Dyer, C. 2014.
\newblock Improving vector space word representations using multilingual
  correlation.
\newblock In \emph{European Chapter of the Association for Computational
  Linguistics}.

\bibitem[{Hinton, Vinyals, and Dean(2015)}]{hinton2015distilling}
Hinton, G.; Vinyals, O.; and Dean, J. 2015.
\newblock Distilling the knowledge in a neural network.
\newblock \emph{arXiv preprint arXiv:1503.02531} .

\bibitem[{Hu et~al.(2020)Hu, Ruder, Siddhant, Neubig, Firat, and
  Johnson}]{hu2020xtreme}
Hu, J.; Ruder, S.; Siddhant, A.; Neubig, G.; Firat, O.; and Johnson, M. 2020.
\newblock Xtreme: A massively multilingual multi-task benchmark for evaluating
  cross-lingual generalization.
\newblock In \emph{International Conference on Machine Learning}.

\bibitem[{Huang et~al.(2019)Huang, Liang, Duan, Gong, Shou, Jiang, and
  Zhou}]{huang2019unicoder}
Huang, H.; Liang, Y.; Duan, N.; Gong, M.; Shou, L.; Jiang, D.; and Zhou, M.
  2019.
\newblock Unicoder: A universal language encoder by pre-training with multiple
  cross-lingual tasks.
\newblock In \emph{Empirical Methods in Natural Language Processing}.

\bibitem[{Huang, Ji, and May(2019)}]{huang2019cross}
Huang, L.; Ji, H.; and May, J. 2019.
\newblock Cross-lingual multi-level adversarial transfer to enhance
  low-resource name tagging.
\newblock In \emph{North American Chapter of the Association for Computational
  Linguistics}.

\bibitem[{Klementiev, Titov, and Bhattarai(2012)}]{klementiev2012inducing}
Klementiev, A.; Titov, I.; and Bhattarai, B. 2012.
\newblock Inducing crosslingual distributed representations of words.
\newblock In \emph{International Conference on Computational Linguistics}.

\bibitem[{Lample and Conneau(2019)}]{lample2019cross}
Lample, G.; and Conneau, A. 2019.
\newblock Cross-lingual language model pretraining.
\newblock In \emph{Advances in Neural Information Processing Systems}.

\bibitem[{Lewis et~al.(2020)Lewis, O{\u{g}}uz, Rinott, Riedel, and
  Schwenk}]{lewis2019mlqa}
Lewis, P.; O{\u{g}}uz, B.; Rinott, R.; Riedel, S.; and Schwenk, H. 2020.
\newblock MLQA: Evaluating cross-lingual extractive question answering.
\newblock In \emph{Association for Computational Linguistics}.

\bibitem[{Liang et~al.(2020)Liang, Duan, Gong, Wu, Guo, Qi, Gong, Shou, Jiang,
  Cao et~al.}]{liang2020xglue}
Liang, Y.; Duan, N.; Gong, Y.; Wu, N.; Guo, F.; Qi, W.; Gong, M.; Shou, L.;
  Jiang, D.; Cao, G.; et~al. 2020.
\newblock Xglue: A new benchmark dataset for cross-lingual pre-training,
  understanding and generation.
\newblock \emph{arXiv preprint arXiv:2004.01401} .

\bibitem[{Mikolov, Le, and Sutskever(2013)}]{mikolov2013exploiting}
Mikolov, T.; Le, Q.~V.; and Sutskever, I. 2013.
\newblock Exploiting similarities among languages for machine translation.
\newblock \emph{arXiv preprint arXiv:1309.4168} .

\bibitem[{Nivre et~al.(2016)Nivre, De~Marneffe, Ginter, Goldberg, Hajic,
  Manning, McDonald, Petrov, Pyysalo, Silveira et~al.}]{nivre2016universal}
Nivre, J.; De~Marneffe, M.-C.; Ginter, F.; Goldberg, Y.; Hajic, J.; Manning,
  C.~D.; McDonald, R.; Petrov, S.; Pyysalo, S.; Silveira, N.; et~al. 2016.
\newblock Universal dependencies v1: A multilingual treebank collection.
\newblock In \emph{International Conference on Language Resources and
  Evaluation}.

\bibitem[{Pan et~al.(2017)Pan, Zhang, May, Nothman, Knight, and
  Ji}]{pan2017cross}
Pan, X.; Zhang, B.; May, J.; Nothman, J.; Knight, K.; and Ji, H. 2017.
\newblock Cross-lingual name tagging and linking for 282 languages.
\newblock In \emph{Association for Computational Linguistics}.

\bibitem[{Phang et~al.(2020)Phang, Htut, Pruksachatkun, Liu, Vania, Kann,
  Calixto, and Bowman}]{phang2020english}
Phang, J.; Htut, P.~M.; Pruksachatkun, Y.; Liu, H.; Vania, C.; Kann, K.;
  Calixto, I.; and Bowman, S.~R. 2020.
\newblock English Intermediate-Task Training Improves Zero-Shot Cross-Lingual
  Transfer Too.
\newblock \emph{arXiv preprint arXiv:2005.13013} .

\bibitem[{Sasaki et~al.(2018)Sasaki, Sun, Schamoni, Duh, and
  Inui}]{sasaki2018cross}
Sasaki, S.; Sun, S.; Schamoni, S.; Duh, K.; and Inui, K. 2018.
\newblock Cross-lingual learning-to-rank with shared representations.
\newblock In \emph{North American Chapter of the Association for Computational
  Linguistics}.

\bibitem[{Siddhant et~al.(2020)Siddhant, Johnson, Tsai, Ari, Riesa, Bapna,
  Firat, and Raman}]{siddhant2020evaluating}
Siddhant, A.; Johnson, M.; Tsai, H.; Ari, N.; Riesa, J.; Bapna, A.; Firat, O.;
  and Raman, K. 2020.
\newblock Evaluating the Cross-Lingual Effectiveness of Massively Multilingual
  Neural Machine Translation.
\newblock In \emph{AAAI}, 8854--8861.

\bibitem[{Tjong Kim~Sang(2002)}]{tjong-kim-sang-2002-introduction}
Tjong Kim~Sang, E.~F. 2002.
\newblock Introduction to the {C}o{NLL}-2002 Shared Task: Language-Independent
  Named Entity Recognition.
\newblock In \emph{{COLING}-02: The 6th Conference on Natural Language Learning
  2002 ({C}o{NLL}-2002)}.

\bibitem[{Tjong Kim~Sang and
  De~Meulder(2003)}]{tjong-kim-sang-de-meulder-2003-introduction}
Tjong Kim~Sang, E.~F.; and De~Meulder, F. 2003.
\newblock Introduction to the {C}o{NLL}-2003 Shared Task: Language-Independent
  Named Entity Recognition.
\newblock In \emph{Proceedings of the Seventh Conference on Natural Language
  Learning at {HLT}-{NAACL} 2003}.

\bibitem[{Wolf et~al.(2019)Wolf, Debut, Sanh, Chaumond, Delangue, Moi, Cistac,
  Rault, Louf, Funtowicz, Davison, Shleifer, von Platen, Ma, Jernite, Plu, Xu,
  Scao, Gugger, Drame, Lhoest, and Rush}]{Wolf2019HuggingFacesTS}
Wolf, T.; Debut, L.; Sanh, V.; Chaumond, J.; Delangue, C.; Moi, A.; Cistac, P.;
  Rault, T.; Louf, R.; Funtowicz, M.; Davison, J.; Shleifer, S.; von Platen,
  P.; Ma, C.; Jernite, Y.; Plu, J.; Xu, C.; Scao, T.~L.; Gugger, S.; Drame, M.;
  Lhoest, Q.; and Rush, A.~M. 2019.
\newblock HuggingFace's Transformers: State-of-the-art Natural Language
  Processing.
\newblock \emph{arXiv preprint arXiv:1910.03771} .

\bibitem[{Wu et~al.(2020)Wu, Lin, Karlsson, Lou, and Huang}]{wu2020single}
Wu, Q.; Lin, Z.; Karlsson, B.~F.; Lou, J.-G.; and Huang, B. 2020.
\newblock Single-/Multi-Source Cross-Lingual NER via Teacher-Student Learning
  on Unlabeled Data in Target Language.
\newblock In \emph{Association for Computational Linguistics}.

\bibitem[{Xu et~al.(2018)Xu, Yang, Otani, and Wu}]{xu2018unsupervised}
Xu, R.; Yang, Y.; Otani, N.; and Wu, Y. 2018.
\newblock Unsupervised cross-lingual transfer of word embedding spaces.
\newblock In \emph{Empirical Methods in Natural Language Processing}.

\bibitem[{Yang et~al.(2019)Yang, Zhang, Tar, and Baldridge}]{yang2019paws}
Yang, Y.; Zhang, Y.; Tar, C.; and Baldridge, J. 2019.
\newblock PAWS-X: A cross-lingual adversarial dataset for paraphrase
  identification.
\newblock In \emph{Empirical Methods in Natural Language Processing}.

\bibitem[{Zhu et~al.(2019)Zhu, Wang, Wang, Zhou, Zhang, Wang, and
  Zong}]{zhu2019ncls}
Zhu, J.; Wang, Q.; Wang, Y.; Zhou, Y.; Zhang, J.; Wang, S.; and Zong, C. 2019.
\newblock NCLS: Neural cross-lingual summarization.
\newblock In \emph{Empirical Methods in Natural Language Processing}.

\bibitem[{Zweigenbaum, Sharoff, and Rapp(2018)}]{zweigenbaum2018overview}
Zweigenbaum, P.; Sharoff, S.; and Rapp, R. 2018.
\newblock Overview of the third BUCC shared task: Spotting parallel sentences
  in comparable corpora.
\newblock In \emph{Proceedings of 11th Workshop on Building and Using
  Comparable Corpora}.

\end{thebibliography}
